\documentclass[usletter, 10 pt, conference]{ieeeconf}  

\IEEEoverridecommandlockouts                              

\usepackage[pdftex]{graphicx}
\usepackage{cite}
\usepackage{amsmath}
\usepackage{amsfonts}
\usepackage{array}
\usepackage{url}
\usepackage{subcaption}
\usepackage{xcolor}
\usepackage{tikz}

\usepackage{float}

\title{\LARGE \bfseries
	Training Adversarial Agents to Exploit Weaknesses in Deep Control Policies
}

\author{Sampo Kuutti,
	Saber Fallah, 
	\thanks{Sampo Kuutti and Saber Fallah are with the Connected and Autonomous Vehicles Lab, University of Surrey, Guildford, GU2 7XH, UK. Email: \{s.j.kuutti, s.fallah\}@surrey.ac.uk}
	Richard Bowden
	\thanks{Richard Bowden is with the Centre for Vision, Speech and Signal Processing, University of Surrey, GU2 7XH, UK. Email: r.bowden@surrey.ac.uk}
}

\begin{document}
	
	\maketitle
	\thispagestyle{empty}
	\pagestyle{empty}

	\begin{abstract}
		Deep learning has become an increasingly common technique for various control problems, such as robotic arm manipulation, robot navigation, and autonomous vehicles. However, the downside of using deep neural networks to learn control policies is their opaque nature and the difficulties of validating their safety. As the networks used to obtain state-of-the-art results become increasingly deep and complex, the rules they have learned and how they operate become more challenging to understand. This presents an issue, since in safety-critical applications the safety of the control policy must be ensured to a high confidence level. In this paper, we propose an automated black box testing framework based on adversarial reinforcement learning. The technique uses an adversarial agent, whose goal is to degrade the performance of the target model under test. We test the approach on an autonomous vehicle problem, by training an adversarial reinforcement learning agent, which aims to cause a deep neural network-driven autonomous vehicle to collide. Two neural networks trained for autonomous driving are compared, and the results from the testing are used to compare the robustness of their learned control policies. We show that the proposed framework is able to find weaknesses in both control policies that were not evident during online testing and therefore, demonstrate a significant benefit over manual testing methods.
	\end{abstract}

	\section{Introduction}
		The rise of deep learning has resulted in rapid progress in many fields, with state-of-the-art results obtained in fields such as image classification, sound recognition, and language processing \cite{krizhevsky2012imagenet, hinton2012deep, sutskever2014sequence}. The strong capability of Deep Neural Networks (DNNs) for modelling highly non-linear and complex functions has resulted in the adoption of DNNs in many control problems. Important results in control applications such as robotic arm manipulation, robot navigation, and autonomous vehicle control have been achieved through deep learning \cite{levine2016end, gu2017deep, lee2019making, zhu2017target, kuutti2020survey, bojarski2016end, codevilla2018end}. However, in safety-critical applications, the safety of the control policy must be fully guaranteed before it is commercially deployable. This presents a significant obstacle to the deployment of DNN-based control policies in safety-critical applications such as autonomous driving \cite{varshney2017safety, borg2018safely}. As the operational environment of the system becomes increasingly complex, it becomes infeasible to test the control policy in all possible scenarios it may encounter \cite{burton2017making, kalra2016driving, wachenfeld2017new, coelingh2018driving}. Therefore, methods for testing and understanding the safety of these opaque systems are necessary \cite{koopman2016challenges, van2017challenges, castelvecchi2016can, adadi2018peeking}. Moreover, in tasks such as autonomous driving, testing the system in a naturalistic driving environment would mean that edge cases where collisions are more likely to occur, would be seen rarely \cite{codevilla2019exploring}. Therefore, by using an adversarial agent whose aim is to deliberately create these edge case scenarios, better insights into possible failure cases can be obtained with reduced training times.  
		
		The concept of utilising an adversarial agent to disturb a machine learning agent has been suggested previously, for example, by Morimoto \& Doya \cite{morimoto2005robust}, who used an actor-disturbor-critic method, where the disturbor aimed to find the worst disturbance to reduce the performance of a controller. This was used in the training loop of a reinforcement learning agent to improve the robustness of the control policy to disturbances, and was demonstrated in an inverted pendulum task. The framework was extended to use DNNs for estimating the control policy and disturbances in a deep reinforcement learning framework by Pinto et al. \cite{pinto2017robust}, and was demonstrated successfully in a robotic manipulation task. For autonomous vehicles, the idea of learning to automatically find failure cases was suggested as early as 1992, by Schultz et al. \cite{schultz1992adaptive}, who used genetic algorithms to find test cases that exposed weaknesses in autonomous aerial vehicle controllers. The results suggested this could be an effective alternative to manual testing of complex software controllers. In more recent work, Behzadan \& Munir \cite{behzadan2018adversarial} demonstrated that a reinforcement learning agent could be trained to create collisions with other road vehicles, by training an agent to collide against two agents, a DNN and a rule-based system. The number of episodes to convergence and minimum time-to-collision were then used to argue the DNN was the safer control policy. However, by having no constraints on the adversarial agent it is likely to learn a behaviour unlike any human driver, which could limit insights into plausible collision cases that might happen if the DNN control policies were deployed in the real world. For instance, in the examples shown by Behzadan \& Munir \cite{behzadan2018adversarial}, the adversarial agent approached the target vehicle from the rear at high velocity, making collision avoidance extremely difficult. Moreover, this type of collision does not necessarily represent a vulnerability in the control policy under test, as the adversarial agent would be considered at fault in a real world collision \cite{shalev2017formal}. Perhaps the closest work to our research is Adaptive Stress Testing (AST) by Koren et al. \cite{koren2018adaptive}. AST aims to find the most likely collision cases for an autonomous vehicle by manipulating the actions of pedestrians in the simulation environment and the noise in the observations of the control policy under testing. However, this approach has several weaknesses which limit the insight it can offer into the vulnerabilities in the autonomous system under testing. For example, in majority of the collisions found, the blame for the collision would fall on the pedestrians controlled by AST. Furthermore, the AST framework was only evaluated on a simple rule-based vehicle following system. Instead, in our approach there are constraints on the behaviour of the adversarial agent to maintain plausible driving trajectories and the focus is to find vulnerabilities which lead to collisions where the autonomous vehicle being tested is at fault. Moreover, the observations of the system under testing are not manipulated in any way, therefore all collision cases found by the proposed framework demonstrate a vulnerability in the learned deep control policy.
		
		In this paper, we propose a technique for targeted black box testing, using a reinforcement learning algorithm to find the test scenarios which are most likely to cause the black box control policy to fail. The proposed system has no knowledge of the internal mechanisms of the control policy under testing, but instead learns a behaviour which finds failure cases for the control policy. In this way, the powerful function approximation capabilities of DNNs are used to find the weaknesses in other DNNs, and the testing procedure can therefore be fully automated. The proposed framework is tested in an autonomous driving problem, where the Adversarial Reinforcement Learning (ARL) agent is attempting to cause a vehicle following model to crash. Note that our approach is distinct to work on adversarial attacks \cite{szegedy2013intriguing, goodfellow2014explaining, papernot2016limitations}, as we are not manipulating the inputs to the target DNN, instead we place another agent in the same environment which aims to deliberately cause the target control policy to fail. Similarly, our approach is distinct to research into adversarial robustness \cite{xiao2018training, tsipras2018robustness, schmidt2018adversarially}, as we do not aim to train the model to be robust to adversarial examples, instead we aim to leverage the adversarial agent to find failure cases in the target models more reliably than manual testing methods can, and understand the weaknesses present in the deep control policies.
		
		The remainder of this paper is structured as follows. Section \ref{sec_arl} presents the necessary background, methodology, and general framework behind ARL. The simulations results of the vehicle following use case are presented in Section \ref{sec_sim}. Finally, concluding remarks are given in Section \ref{sec_conc}.
	
	\section{Methodology} \label{sec_arl}
		\subsection{Markov Decision Processes}
		Reinforcement learning allows an agent to learn through interaction with its environment. Reinforcement learning can be formally described by a Markov Decision Process (MDP), denoted by a tuple \{$\mathcal{S}, \mathcal{A}, \mathcal{P}, \mathcal{R}$\}, where $\mathcal{S}$ represents the state space, $\mathcal{A}$ represents the action space, $\mathcal{P}$ denotes the state transition probability model, and $\mathcal{R}$ is the reward function. At each time step $t$, the agent observes state $s_t \in \mathcal{S}$ and takes an action $a_t \in \mathcal{A}$, according to its policy $\pi(s_t)$, causing the environment to transition to the next state $s_{t+1}$ according to the transition dynamics $p(s_{t+1}|s_t, a_t)$ as given by the transition probability model $\mathcal{P}$. The agent then receives a reward $r_t$, according to the reward function $\mathcal{R}$, and observes the new state of the environment $s_{t+1}$. The network parameters are then updated, such that the expected future rewards are maximised. As the agent interacts with the environment it learns through trial-and-error a state-action mapping for an optimal policy $\pi^*(s_t)$, which maximises the discounted sum of rewards over time given by the returns $R_t$. Therefore, this exploration of the operational environment can be leveraged to explore potential weaknesses in black box systems.
		\begin{equation}
		R_t = \sum_{k=0}^{\infty}\gamma^kr_{t+k}
		\end{equation}
		where $\gamma \in [0,1]$ is the discount factor used to prioritise immediate rewards over future rewards.
		
		\subsection{Reinforcement Learning}
\begin{figure}[h]
			\centering
			\includegraphics[width=0.45\textwidth]{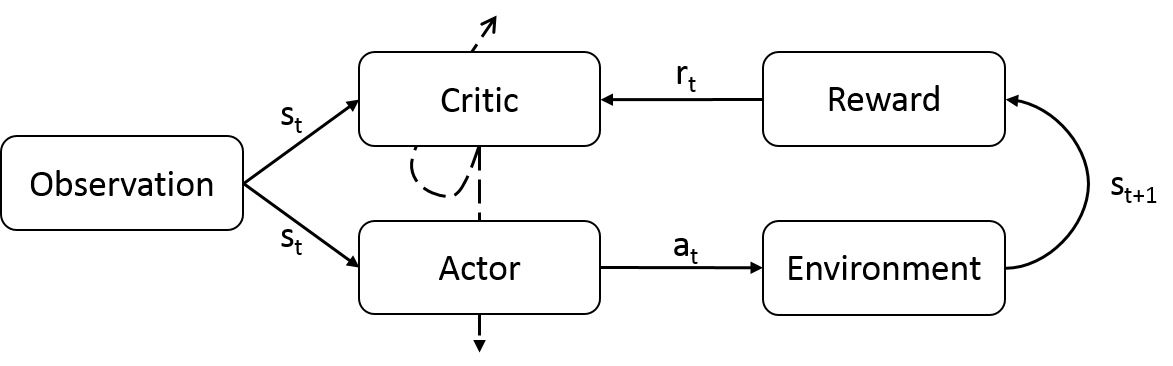}
			\caption{An actor-critic network architecture. The dashed lines represent network updates\cite{kuutti2019end}.}
			\label{fig_ac}
		\end{figure}

		In our framework, the algorithm used to train the adversarial agent is Advantage Actor Critic (A2C) \cite{mnih2016asynchronous}, which uses an actor-critic network architecture, as shown in Fig. \ref{fig_ac}. The actor network estimates the optimal policy function $\pi^*(s_t)$, which aims to maximise the expected rewards. Meanwhile, the critic network estimates the value of being in a given state, with the Value function $V(s)$. The weights of both networks are then updated based on the Advantage function $A(s_t, a_t)$:
		\begin{equation}
		V(s) = \mathbb{E}[R_t|s_t=s]
		\end{equation}
		\begin{equation}
		Q(s,a) = \mathbb{E}[R_t|s_t=s, a]
		\end{equation}
		\begin{multline}
			A(s_t, a_t) = Q(s_t, a_t) - V (s_t) \\
			\approx \sum_{k}^{n-1} \gamma^kr_{t+k} + \gamma^n V(s_{t+n}) - V(s_t)
		\end{multline}
		Where $\mathbb{E}$ denotes expectation, $V(s_t)$ is the value function, and $Q(s_t, a_t)$ is the quality function estimating the value of each action for a given state \cite{bhatnagar2008incremental, sutton1998reinforcement}.
		
		The network architectures for both networks are as follows. The actor network has 3 fully-connected layers, followed by a Long Short-Term Memory (LSTM) \cite{hochreiter1997long} layer, which is fully connected to the output layer. The actor network estimates the stochastic control policy with two outputs, mean value $\mu$ and estimated variance $\sigma^2$, which are used to generate a Gaussian distribution $\mathcal{N}$ from which the action is sampled, such that $a\sim\mathcal{N}(\mu, \sigma^2)$. Meanwhile, the critic network uses only 2 fully-connected layers followed by the output layer to estimate the value function $V(s)$. All hidden neurons use a ReLU-6 activation \cite{krizhevsky2010convolutional}, whilst the $\mu$ uses a tanh activation, the $\sigma^2$ uses a softplus activation, and the value estimate has a linear activation. 
		
		A2C training is formulated as in \cite{kuutti2019end}, by updating the actor and critic networks in separate update steps, using a policy loss $\mathcal{L}_{\pi}$ and value loss $\mathcal{L}_v$ functions, respectively, as given by:
		\begin{equation}
			\mathcal{L}_{v} = (A(s_t, a_t))^2
		\end{equation}
		\begin{equation}
			\mathcal{L}_{\pi} = -log\pi(a_t| s_t)A(s_t,a_t) - \beta H(\pi(s_t))
		\end{equation}
		where $\beta$ is the entropy coefficient and $H(\pi(s_t))$ is the entropy added to encourage exploration in the policy, calculated as
		\begin{equation}
			H(\pi(s_t)) = \frac{1}{2}(log(2\pi \sigma^2)+1)
		\end{equation}
		
		Both networks are updated using RMSProp optimiser \cite{tieleman2012lecture} during training, using their respective loss functions. The final hyperparameters of the network architecture are shown in Table \ref{table_netarch}.
		
		\begin{table}
			\caption{Final network hyperparameters.}
			\label{table_netarch}
			\centering
			\begin{tabular}{ m{5cm}  m{1cm} }
				\hline
				\hline
				\textbf{Parameter} & \textbf{Value} \\
				\hline
				No. hidden layers (actor) & 3 \\
				No. neurons per hidden layer (actor) & 50 \\
				No. of LSTM units (actor) & 16 \\
				No. hidden layers (critic) & 2 \\
				No. neurons per hidden layer (critic) & 50 \\
				Learning rate (actor), \textit{$\eta_{actor}$} & 1x10\textsuperscript{-4} \\
				Learning rate (critic), \textit{$\eta_{critic}$} & 1x10\textsuperscript{-2} \\
				Discount factor, \textit{$\gamma$} & 0.99 \\
				Entropy coefficient, \textit{$\beta$} & 1x10\textsuperscript{-4} \\
				RMSProp $\epsilon$ & 1x10\textsuperscript{-10} \\
				RMSProp decay $\alpha$ & 0.9 \\
				RMSProp momentum & 0.0 \\
				\hline
				\hline
			\end{tabular}
		\end{table}

	\subsection{Training Environment}
		The autonomous driving simulation was defined as a vehicle following scenario in highway driving. Two vehicles are driving at highway speeds on a straight road. The follower is a DNN trained to follow a leading vehicle at a safe distance, whilst the lead vehicle is the adversarial agent whose aim is to find weaknesses in the follower's control policy. In order to do this, the adversarial agent must create collisions, thus proving the follower's control policy is unsafe. For this scenario, the input to the ARL network are the follower vehicle velocity $v_{f}$, follower vehicle acceleration $a_{f}$, relative velocity to the follower $v_{rel}$, and time headway between the two vehicles $t_h$, such that $s_t^{ARL} = [v_f, a_f, v_{rel}, t_h]$. The output of the network is the lead vehicle acceleration for the next time step. The simulation time steps are fixed at 40 ms.
		
		We demonstrate this framework by attacking two previously published DNN models trained for vehicle following using the IPG CarMaker simulator \cite{IPG2017}, (1) a Reinforcement Learning (RL) model \cite{kuutti2019end} and (2) an Imitation Learning (IL) model \cite{kuutti2019safe}. The RL model uses a feedforward network with an LSTM layer to control the longitudinal actions of the vehicle using the observations $s_{t}^{RL} = [v_f, a_f, v_{rel}, t_h]$. The IL model uses a simple feedforward network to also control the longitudinal actions of the vehicle, using the observations $s_t^{IL} = [v_f, v_{rel}, t_h]$. Both models aim to maintain a 2 s time headway $t_h$ from the lead vehicle. The time headway is a measure of intervehicular distance in time, given as follows:
		\begin{equation} \label{eq_th}
			t_{h} = \frac{x_{rel}}{v_f}
		\end{equation}
		where $x_{rel}$ is the relative distance between the two vehicles in m, and $v_f$ is the velocity of the following vehicle in m/s.
		
		The training was broken down into 5-minute episodes, where the episode ends after the 5 minutes have passed or a collision occurs. At the start of each episode, a road friction coefficient $CoF \in$ \{0.4, 0.425, ... , 1.0\} was randomly chosen. It should be noted that a collision may be easier to cause in low friction conditions as the reaction time required for the follower vehicle reduces \cite{reif2014brakes}, however none of the agents can observe the road friction coefficients and should therefore learn a policy which generalises to different road conditions. The reward function $\mathcal{R}$ for training the ARL agent was given based on the time headway:
		\begin{equation}
			\mathcal{R} = \min{\left(\frac{1}{t_h}, 100\right)}
		\end{equation}
		Thus, the reward function rewards low time headways, encouraging collisions to occur. The reward is capped at 100, as otherwise the reward function would tend towards infinity as the time headway reaches zero.
		
		The velocity and the acceleration of the lead vehicle were limited to ensure that the vehicle behaviour remains plausible and the velocity is in the highway driving range, as well as to obtain insights into the effect of the driving speeds on the robustness of the vehicle following models. The acceleration was always limited to $a_{lead} \in$ [-6, 2] m/s\textsuperscript{2}, whilst four velocity ranges were tested as $v_{lead} \in$ [17, 30], [12, 35], [12, 30], [17, 35] m/s. For each velocity constraint and vehicle follower model combination, 5 training runs of 2,500 episodes were completed.
		
	\section{Simulation Results} \label{sec_sim}
	\subsection{Results}
\begin{figure*}
				\centering
				\begin{subfigure}{0.45\textwidth}
					\includegraphics[width=\textwidth]{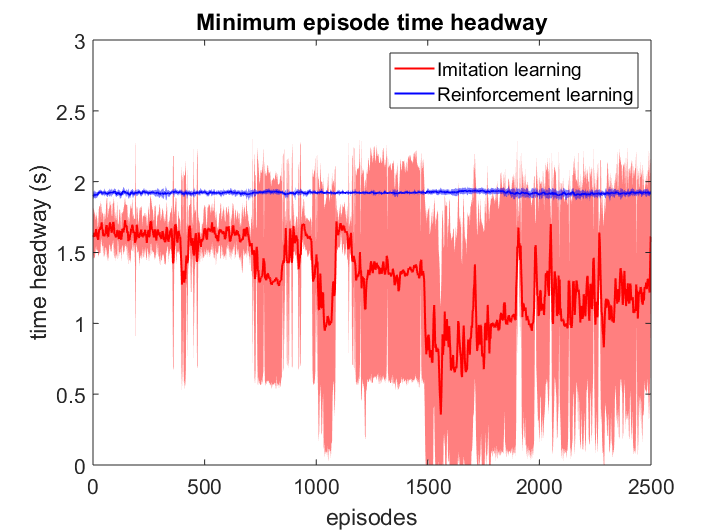}
					\label{subfig_ipg_17-30}
					\caption{$v_{lead} \in$ [17, 30] m/s.}
				\end{subfigure}
				\begin{subfigure}{0.45\textwidth}
					\includegraphics[width=\textwidth]{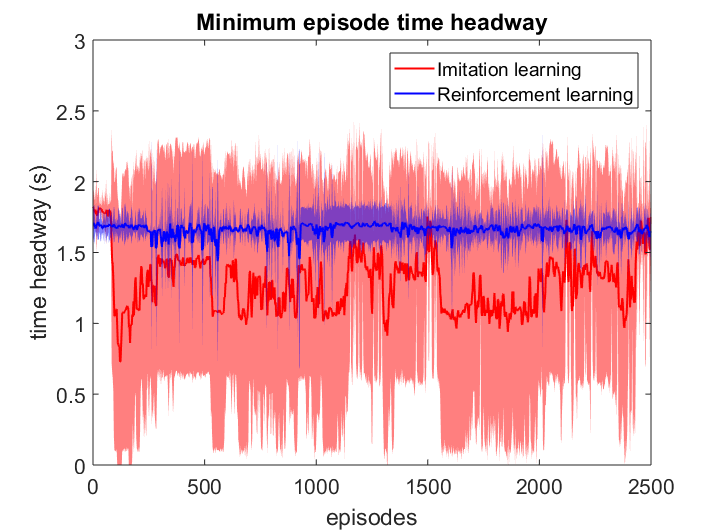}
					\label{subfig_ipg_12-35}
					\caption{$v_{lead} \in$ [12, 35] m/s.}
				\end{subfigure}
				\begin{subfigure}{0.45\textwidth}
					\includegraphics[width=\textwidth]{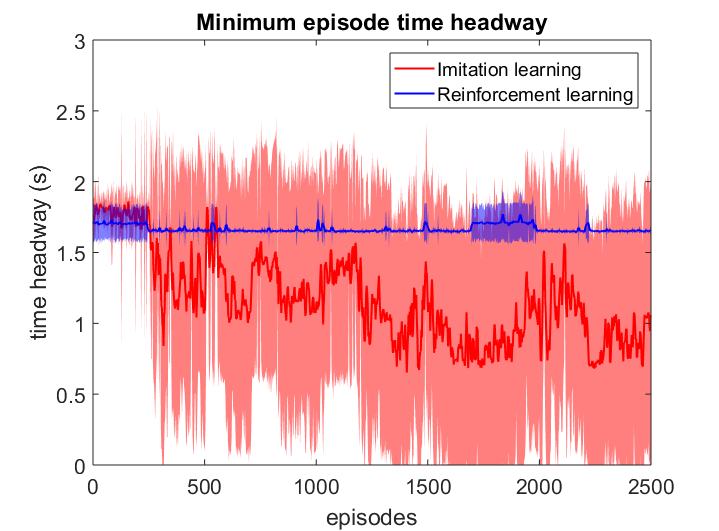}
					\label{subfig_ipg_12-30}
					\caption{$v_{lead} \in$ [12, 30] m/s.}
				\end{subfigure}
				\begin{subfigure}{0.45\textwidth}
					\includegraphics[width=\textwidth]{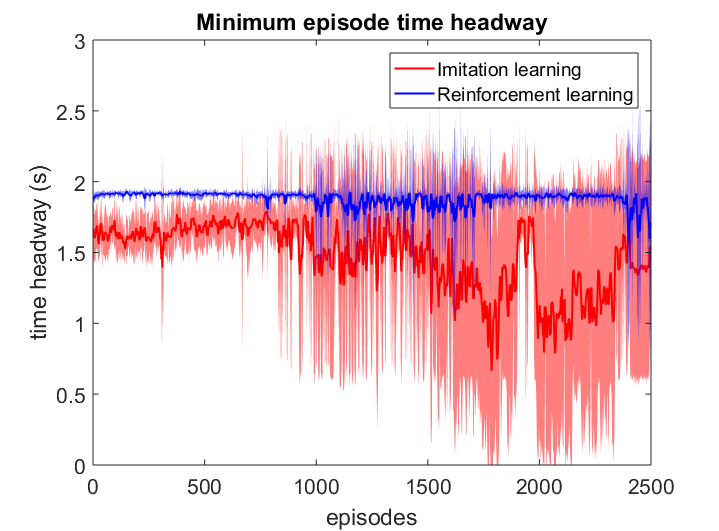}
					\label{subfig_ipg_17-35}
					\caption{$v_{lead} \in$ [17, 35] m/s.}
				\end{subfigure}
				\caption{Comparison of the two vehicle following agents' minimum $t_h$ per episode over training runs. Averaged over 5 runs, with standard deviation shown in shaded colour.}
				\label{fig_ipg}
			\end{figure*}

		The average number of collisions and episodes until first collision for each velocity range and vehicle follower model can be seen in Tables \ref{table_collisions} and \ref{table_eptocollision}, respectively. In initial testing, the lead vehicle was limited to $v_{lead} \in$ [17, 30] m/s. Since the vehicle following models were trained in this velocity range, it tests their robustness in their training domain. The ARL agent was then trained for 2,500 episodes against both agents, for which the results can be seen in Fig. \ref{fig_ipg}(a). The results demonstrate the IL model is susceptible to an adversarial agent, and thus the ARL agent can cause collisions to occur. On the other hand, the RL model has zero collisions with the ARL agent, and as can be seen from Fig. \ref{fig_ipg}(a) the minimum time headway in the episodes remains near the target headway of 2 s. This shows a significant benefit of the RL model over the IL one, in terms of robustness to an adversarial agent. The second set of experiments, shown in Fig. \ref{fig_ipg}(b), relaxed the velocity constraints on the lead vehicle, to $v_{lead} \in$ [12, 35] m/s increasing the maximum velocity and decreasing the minimum velocity. These velocity ranges are outside the distribution the vehicle following models experienced during training, and therefore also test model generalisation capability. From the results, it can be seen that both models are more susceptible to an attack in this domain, but nevertheless the RL model still demonstrates significant safety benefits over the IL model. The two last velocity ranges tested were $v_{lead} \in$ [12, 35] and [17, 30] m/s, relaxing the minimum and maximum lead vehicle velocity constraints, respectively. The results can be seen in Fig. \ref{fig_ipg}(c) and (d). Comparing the two sets of experiments, it can be seen that relaxing the minimum velocity and allowing the lead vehicle to drive at lower speeds enables it to find collision cases more easily. In both cases, collision cases against the IL model are found. However, the results from Tables \ref{table_collisions} and \ref{table_eptocollision} show that the ARL is able to exploit the IL model significantly more often and earlier in its training. On the other hand, the RL model only collides in the higher velocity experiments, although this occurs relatively rarely and only at the very end of the ARL agent's training phase.
		
			\begin{table}
				\renewcommand{\arraystretch}{1.3}
				\caption{Average number of collisions for different lead vehicle velocity constraints. Averaged over 5 training runs of 2,500 episodes each.}
				\label{table_collisions}
				\centering
				\begin{tabular}{c c c}
					\hline
					\hline
					\textbf{v\textsubscript{lead}} & \textbf{Imitation Learning} & \textbf{Reinforcement Learning} \\ 
					\hline
					\textbf{[17, 30] m/s} & 486.6 & 0.0 \\
					\textbf{[12, 35] m/s} & 644.0 & 2.0 \\
					\textbf{[12, 30] m/s} & 799.6  & 0.0  \\
					\textbf{[17, 35] m/s} & 315.2  & 1.2 \\
					\hline
					\hline
					
				\end{tabular}
			\end{table}
			
			\begin{table}
				\renewcommand{\arraystretch}{1.3}
				\caption{Average number of episodes until first collision found for different lead vehicle velocity constraints. Averaged over 5 training runs of 2,500 episodes each.}
				\label{table_eptocollision}
				\centering
				\begin{tabular}{c c c}
					\hline
					\hline
					\textbf{v\textsubscript{lead}} & \textbf{Imitation Learning} & \textbf{Reinforcement Learning} \\ 
					\hline
					\textbf{[17, 30] m/s} & 563.2 & 0.0 \\
					\textbf{[12, 35] m/s} & 579.2 & 922.3 \\
					\textbf{[12, 30] m/s} & 245.3  & 0.0 \\
					\textbf{[17, 35] m/s} & 1030.8 & 2451.0 \\
					\hline
					\hline
					
				\end{tabular}
			\end{table}
		
\begin{figure*}
			\centering
			\begin{subfigure}{0.45\textwidth}
				\includegraphics[width=\textwidth]{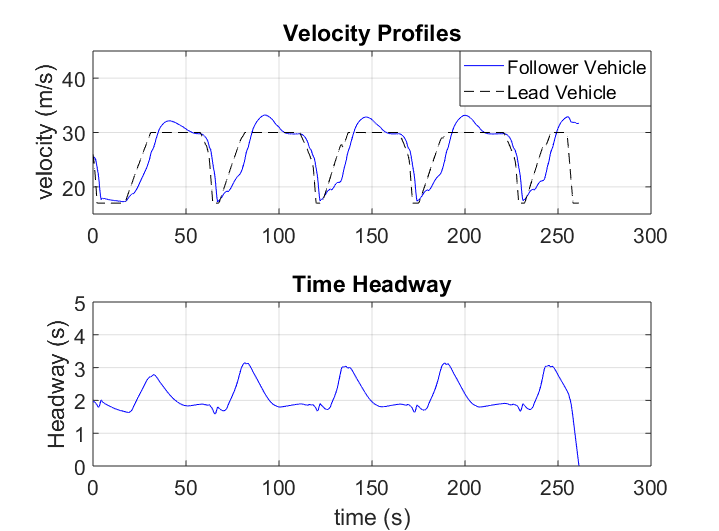}
				\label{subfig_collisions_p5_400}
				\caption{}
			\end{subfigure}
			\begin{subfigure}{0.45\textwidth}
				\includegraphics[width=\textwidth]{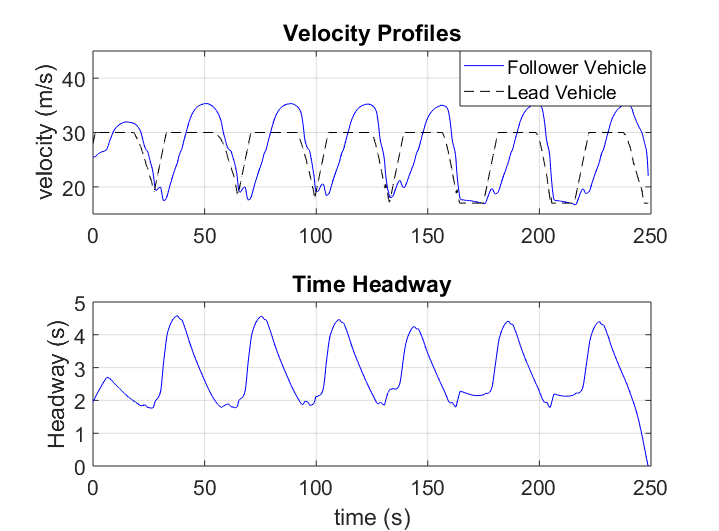}
				\label{subfig_collisions_p5_1325}
				\caption{}
			\end{subfigure}
			\begin{subfigure}{0.45\textwidth}
				\includegraphics[width=\textwidth]{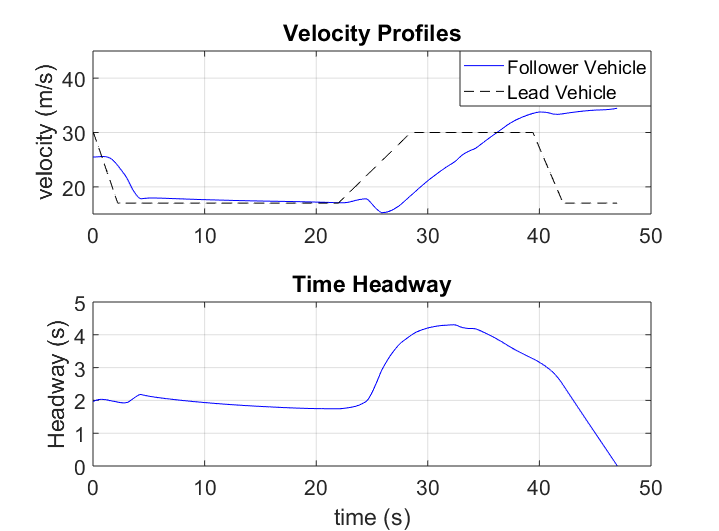}
				\label{subfig_collisions_p6_1631}
				\caption{}
			\end{subfigure}
			\begin{subfigure}{0.45\textwidth}
				\includegraphics[width=\textwidth]{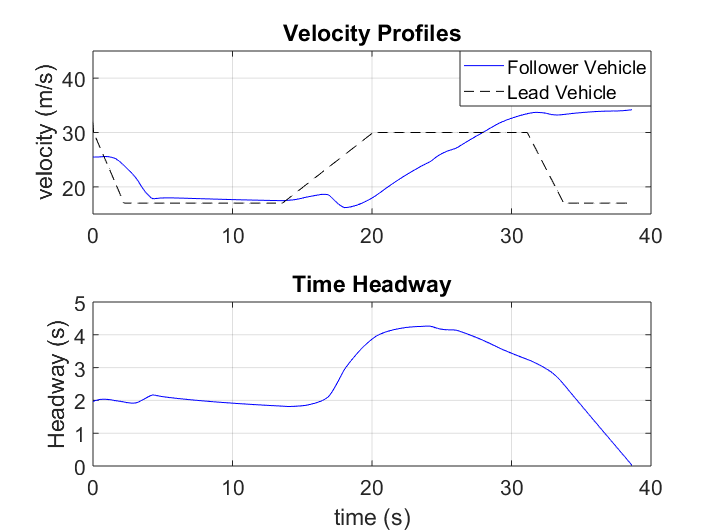}
				\label{subfig_collisions_p6_1633}
				\caption{}
			\end{subfigure}
			\caption{Comparison of collision scenarios between training runs, (a) and (b) are from training run 1, whilst (c) and (d) are from training run 2. Both training runs use velocity constraints of $v_{lead} \in$ [17, 30] m/s and the IL model as the vehicle follower.}
			\label{fig_collisions}
		\end{figure*}

		Further investigation into the type of behaviour the ARL was adopting during training revealed that, for a single training run, the ARL tends to converge to a singular type of behaviour that leads to collisions and these behaviours can vary significantly between different training runs. While some differences in the converged behaviour of the agent can be expected due to the variance in reinforcement learning \cite{henderson2018deep, schulman2015high, romoff2018reward, weaver2001optimal}, these results show significant differences between different trained agents. For instance, example collision scenarios are shown in Fig \ref{fig_collisions}, where 2 collisions from 1 training run are shown in the top subfigures, whilst 2 collisions from another training run are shown in the bottom subfigures. For consistency, both training runs are attacking the IL model, with the same velocity constraints. As can be seen in the first two plots, the ARL agent has adopted a strategy in which it continuously accelerates and decelerates between high and low velocities, until the follower vehicle comes close to it with a high acceleration rate, at which point the lead vehicle then decelerates at maximum deceleration. Meanwhile, in the plots (c) and (d), the ARL agent has adopted a strategy in which it first decelerates to a low velocity, and once both vehicles are at low velocities it starts to accelerate back to the maximum velocity, followed by waiting until the following vehicle is approaching it at high acceleration, when it finally decelerates and creates a collision. These results reveal a flaw in the IL model, where it continues to accelerate when the $t_h >$ 2 s, trying to reach the target $t_h$ of 2 s, even if the lead vehicle is decelerating and there is a large relative velocity difference between the vehicles. Finding different collision modes is beneficial, as it offers further insight into the different vulnerabilities present in the control policy. Therefore, by exploiting information from multiple training runs where the ARL is using different collision modes, valuable insight into the weaknesses of the DNN under testing can be obtained.
		
		\subsection{Discussion}
			The overall testing completed accounts for a total of 100,000 episodes, or over 8000 simulated hours of testing. This resulted in a total of 11243 collision cases found, which includes 11227 and 16 for the IL and RL models, respectively. This clearly demonstrates the significantly higher robustness of the RL model to the presence of an adversarial agent. Moreover, these results demonstrate that the proposed ARL framework is able to find failure cases for both control policies under testing. Compared to the type of manual test case definition often used for vehicle safety testing, this can be highly beneficial for testing complex black box control systems. For instance, both control policies tested here, were tested for 10 hours of simulated vehicle following in their original works, where the lead vehicle also drove at highway speeds. In this manual testing, the types of trajectories executed by the lead vehicle were manually defined (including both naturalistic driving and emergency manoeuvrers), where the parameters (e.g. maximum velocity, acceleration, time to execute manoeuvrer etc.) were randomised during testing. The constraints on the lead vehicle used in the manual test case definition were $v_{lead} \in [17,40]$ m/s and $a_{lead} \in [-6, 2]$ m/s\textsuperscript{2}, and road friction coefficient was uniformly sampled from $CoF \in [0.4, 1.0]$, representing similar driving conditions to those in the adversarial testing framework presented here. The results for these driving tests are shown in Table \ref{table_simresults} and show that during normal testing not a single collision was found. This demonstrates how effective our ARL is at finding weaknesses in DNN-based control policies. Indeed, the results from the manual testing would suggest the IL model to be the safer control policy. However, our testing framework exposes significant vulnerabilities in the IL model, demonstrating that the RL control policy is significantly more robust to the presence of an adversarial agent.
			
				\begin{table}
				\renewcommand{\arraystretch}{1.3}
				\caption{10-hour driving test with manually defined lead vehicle trajectories.}
				\label{table_simresults}
				\centering
				\begin{tabular}{ c c c }
					\hline
					\hline
					\textbf{Parameter} & \textbf{Imitation Learning} & \textbf{Reinforcement Learning} \\
					\hline
					min. x\textsubscript{rel} & 23.844 m & 7.780 m \\
					mean x\textsubscript{rel} & 57.37 m & 58.01 m \\
					max. v\textsubscript{rel} & 8.878 m/s & 7.891 m/s \\
					mean v\textsubscript{rel} & 0.0197 m/s & 0.0289 m/s \\
					min. t\textsubscript{h} & 1.738 s & 1.114 s \\
					mean t\textsubscript{h}& 1.990 s & 2.007 s \\
					collisions & 0 & 0 \\ 
					\hline
					\hline
				\end{tabular}
			\end{table}
	
	\section{Concluding Remarks} \label{sec_conc}
		In this paper, an automated testing framework for deep neural networks was presented. The proposed framework is based on adversarial reinforcement learning, where an adversarial agent is placed in the same environment with the system under testing. By training the adversarial agent through reinforcement learning, the agent learns behaviours which degrade the performance of the target system. This general concept could be used to analyse vulnerabilities in control policies used in multi-agent environments, such as robotic manipulation or unmanned aerial vehicles. In our work, the ARL approach was tested in an autonomous vehicle use case, where the aim of the ARL agent was to cause the vehicle behind it to collide into it. Two neural network models trained for vehicle following were tested, one which uses imitation learning and the other using reinforcement learning. Both models had no collisions when manually tested in their original works. The ARL agent was shown to be able to learn a driving behaviour which can cause both target models to collide into the lead vehicle. This in itself demonstrates the significant benefit of this type of targeted adversarial black box testing. Also, the results showed that the reinforcement learning model is significantly more robust to this kind of adversarial behaviour, demonstrating the safety benefit of the reinforcement learning model over the imitation learning model. This type of adversarial testing framework provides an important technique for testing black box control policies, and can be used to benchmark and compare deep control policies as well as to gain additional insights into the types of edge cases the policies are likely to fail in.
		
		\addtolength{\textheight}{-4.0cm}   

	\section*{Acknowledgment}
	\addcontentsline{toc}{section}{Acknowledgment}
	This work was funded by the EPSRC under grant agreements (EP/R512217/1) and Innovate UK Autonomous Valet Parking Project (Grant No 104273). We would also like to thank NVIDIA Corporation for their GPU grant.
	
	\bibliographystyle{IEEEtran}
	\bibliography{arl}

\begin{thebibliography}{10}
\providecommand{\url}[1]{#1}
\csname url@samestyle\endcsname
\providecommand{\newblock}{\relax}
\providecommand{\bibinfo}[2]{#2}
\providecommand{\BIBentrySTDinterwordspacing}{\spaceskip=0pt\relax}
\providecommand{\BIBentryALTinterwordstretchfactor}{4}
\providecommand{\BIBentryALTinterwordspacing}{\spaceskip=\fontdimen2\font plus
\BIBentryALTinterwordstretchfactor\fontdimen3\font minus
  \fontdimen4\font\relax}
\providecommand{\BIBforeignlanguage}[2]{{%
\expandafter\ifx\csname l@#1\endcsname\relax
\typeout{** WARNING: IEEEtran.bst: No hyphenation pattern has been}%
\typeout{** loaded for the language `#1'. Using the pattern for}%
\typeout{** the default language instead.}%
\else
\language=\csname l@#1\endcsname
\fi
#2}}
\providecommand{\BIBdecl}{\relax}
\BIBdecl

\bibitem{krizhevsky2012imagenet}
A.~Krizhevsky, I.~Sutskever, and G.~E. Hinton, ``Imagenet classification with
  deep convolutional neural networks,'' in \emph{Advances in Neural Information
  Processing Systems (NIPS)}, 2012, pp. 1097--1105.

\bibitem{hinton2012deep}
G.~Hinton, L.~Deng, D.~Yu, G.~E. Dahl, A.-r. Mohamed, N.~Jaitly, A.~Senior,
  V.~Vanhoucke, P.~Nguyen, T.~N. Sainath \emph{et~al.}, ``Deep neural networks
  for acoustic modeling in speech recognition: The shared views of four
  research groups,'' \emph{IEEE Signal Processing Magazine}, vol.~29, no.~6,
  pp. 82--97, 2012.

\bibitem{sutskever2014sequence}
I.~Sutskever, O.~Vinyals, and Q.~V. Le, ``Sequence to sequence learning with
  neural networks,'' in \emph{Advances in Neural Information Processing Systems
  (NIPS)}, 2014, pp. 3104--3112.

\bibitem{levine2016end}
S.~Levine, C.~Finn, T.~Darrell, and P.~Abbeel, ``End-to-end training of deep
  visuomotor policies,'' \emph{The Journal of Machine Learning Research},
  vol.~17, no.~1, pp. 1334--1373, 2016.

\bibitem{gu2017deep}
S.~Gu, E.~Holly, T.~Lillicrap, and S.~Levine, ``Deep reinforcement learning for
  robotic manipulation with asynchronous off-policy updates,'' in \emph{2017
  IEEE International Conference on Robotics and Automation (ICRA)}.\hskip 1em
  plus 0.5em minus 0.4em\relax IEEE, 2017, pp. 3389--3396.

\bibitem{lee2019making}
M.~A. Lee, Y.~Zhu, K.~Srinivasan, P.~Shah, S.~Savarese, L.~Fei-Fei, A.~Garg,
  and J.~Bohg, ``Making sense of vision and touch: Self-supervised learning of
  multimodal representations for contact-rich tasks,'' in \emph{2019
  International Conference on Robotics and Automation (ICRA)}.\hskip 1em plus
  0.5em minus 0.4em\relax IEEE, 2019, pp. 8943--8950.

\bibitem{zhu2017target}
Y.~Zhu, R.~Mottaghi, E.~Kolve, J.~J. Lim, A.~Gupta, L.~Fei-Fei, and A.~Farhadi,
  ``Target-driven visual navigation in indoor scenes using deep reinforcement
  learning,'' in \emph{2017 IEEE International Conference on Robotics and
  Automation (ICRA)}.\hskip 1em plus 0.5em minus 0.4em\relax IEEE, 2017, pp.
  3357--3364.

\bibitem{kuutti2020survey}
S.~Kuutti, R.~Bowden, Y.~Jin, P.~Barber, and S.~Fallah, ``A survey of deep
  learning applications to autonomous vehicle control,'' \emph{IEEE
  Transactions on Intelligent Transportation Systems}, 2020.

\bibitem{bojarski2016end}
M.~Bojarski, D.~Del~Testa, D.~Dworakowski, B.~Firner, B.~Flepp, P.~Goyal, L.~D.
  Jackel, M.~Monfort, U.~Muller, J.~Zhang \emph{et~al.}, ``End to end learning
  for self-driving cars,'' \emph{arXiv preprint arXiv:1604.07316}, 2016.

\bibitem{codevilla2018end}
F.~Codevilla, M.~Miiller, A.~L{\'o}pez, V.~Koltun, and A.~Dosovitskiy,
  ``End-to-end driving via conditional imitation learning,'' in \emph{2018 IEEE
  International Conference on Robotics and Automation (ICRA)}.\hskip 1em plus
  0.5em minus 0.4em\relax IEEE, 2018, pp. 1--9.

\bibitem{varshney2017safety}
K.~R. Varshney and H.~Alemzadeh, ``On the safety of machine learning:
  Cyber-physical systems, decision sciences, and data products,'' \emph{Big
  data}, vol.~5, no.~3, pp. 246--255, 2017.

\bibitem{borg2018safely}
M.~Borg, C.~Englund, K.~Wnuk, B.~Duran, C.~Levandowski, S.~Gao, Y.~Tan,
  H.~Kaijser, H.~L{\"o}nn, and J.~T{\"o}rnqvist, ``Safely entering the deep: A
  review of verification and validation for machine learning and a challenge
  elicitation in the automotive industry,'' \emph{arXiv preprint
  arXiv:1812.05389}, 2018.

\bibitem{burton2017making}
S.~Burton, L.~Gauerhof, and C.~Heinzemann, ``Making the case for safety of
  machine learning in highly automated driving,'' in \emph{International
  Conference on Computer Safety, Reliability, and Security}.\hskip 1em plus
  0.5em minus 0.4em\relax Springer, 2017, pp. 5--16.

\bibitem{kalra2016driving}
N.~Kalra and S.~M. Paddock, ``Driving to safety: How many miles of driving
  would it take to demonstrate autonomous vehicle reliability?''
  \emph{Transportation Research Part A: Policy and Practice}, vol.~94, pp.
  182--193, 2016.

\bibitem{wachenfeld2017new}
W.~Wachenfeld and H.~Winner, ``The new role of road testing for the safety
  validation of automated vehicles,'' in \emph{Automated Driving}.\hskip 1em
  plus 0.5em minus 0.4em\relax Springer, 2017, pp. 419--435.

\bibitem{coelingh2018driving}
E.~Coelingh, J.~Nilsson, and J.~Buffum, ``Driving tests for self-driving
  cars,'' \emph{IEEE Spectrum}, vol.~55, no.~3, pp. 40--45, 2018.

\bibitem{koopman2016challenges}
P.~Koopman and M.~Wagner, ``Challenges in autonomous vehicle testing and
  validation,'' \emph{SAE International Journal of Transportation Safety},
  vol.~4, no.~1, pp. 15--24, 2016.

\bibitem{van2017challenges}
P.~Van~Wesel and A.~E. Goodloe, ``Challenges in the verification of
  reinforcement learning algorithms,'' {Technical report, NASA}, Tech. Rep.,
  2017.

\bibitem{castelvecchi2016can}
D.~Castelvecchi, ``Can we open the black box of ai?'' \emph{Nature News}, vol.
  538, no. 7623, p.~20, 2016.

\bibitem{adadi2018peeking}
A.~Adadi and M.~Berrada, ``Peeking inside the black-box: A survey on
  explainable artificial intelligence (xai),'' \emph{IEEE Access}, vol.~6, pp.
  52\,138--52\,160, 2018.

\bibitem{codevilla2019exploring}
F.~Codevilla, E.~Santana, A.~M. L{\'o}pez, and A.~Gaidon, ``Exploring the
  limitations of behavior cloning for autonomous driving,'' in
  \emph{Proceedings of the IEEE International Conference on Computer Vision
  (ICCV)}, 2019, pp. 9329--9338.

\bibitem{morimoto2005robust}
J.~Morimoto and K.~Doya, ``Robust reinforcement learning,'' \emph{Neural
  computation}, vol.~17, no.~2, pp. 335--359, 2005.

\bibitem{pinto2017robust}
L.~Pinto, J.~Davidson, R.~Sukthankar, and A.~Gupta, ``Robust adversarial
  reinforcement learning,'' in \emph{Proceedings of the 34th International
  Conference on Machine Learning (ICML)}.\hskip 1em plus 0.5em minus
  0.4em\relax JMLR. org, 2017, pp. 2817--2826.

\bibitem{schultz1992adaptive}
A.~C. Schultz, J.~J. Grefenstette, and K.~A. De~Jong, ``Adaptive testing of
  controllers for autonomous vehicles,'' in \emph{Proceedings of the 1992
  Symposium on autonomous underwater vehicle technology}.\hskip 1em plus 0.5em
  minus 0.4em\relax IEEE, 1992, pp. 158--164.

\bibitem{behzadan2018adversarial}
V.~Behzadan and A.~Munir, ``Adversarial reinforcement learning framework for
  benchmarking collision avoidance mechanisms in autonomous vehicles,''
  \emph{arXiv preprint arXiv:1806.01368}, 2018.

\bibitem{shalev2017formal}
S.~Shalev-Shwartz, S.~Shammah, and A.~Shashua, ``On a formal model of safe and
  scalable self-driving cars,'' \emph{arXiv preprint arXiv:1708.06374}, 2017.

\bibitem{koren2018adaptive}
M.~Koren, S.~Alsaif, R.~Lee, and M.~J. Kochenderfer, ``Adaptive stress testing
  for autonomous vehicles,'' in \emph{2018 IEEE Intelligent Vehicles Symposium
  (IV)}.\hskip 1em plus 0.5em minus 0.4em\relax IEEE, 2018, pp. 1--7.

\bibitem{szegedy2013intriguing}
C.~Szegedy, W.~Zaremba, I.~Sutskever, J.~Bruna, D.~Erhan, I.~Goodfellow, and
  R.~Fergus, ``Intriguing properties of neural networks,'' \emph{arXiv preprint
  arXiv:1312.6199}, 2013.

\bibitem{goodfellow2014explaining}
I.~J. Goodfellow, J.~Shlens, and C.~Szegedy, ``Explaining and harnessing
  adversarial examples,'' \emph{arXiv preprint arXiv:1412.6572}, 2014.

\bibitem{papernot2016limitations}
N.~Papernot, P.~McDaniel, S.~Jha, M.~Fredrikson, Z.~B. Celik, and A.~Swami,
  ``The limitations of deep learning in adversarial settings,'' in \emph{2016
  IEEE European Symposium on Security and Privacy (EuroS\&P)}.\hskip 1em plus
  0.5em minus 0.4em\relax IEEE, 2016, pp. 372--387.

\bibitem{xiao2018training}
K.~Y. Xiao, V.~Tjeng, N.~M. Shafiullah, and A.~Madry, ``Training for faster
  adversarial robustness verification via inducing relu stability,''
  \emph{arXiv preprint arXiv:1809.03008}, 2018.

\bibitem{tsipras2018robustness}
D.~Tsipras, S.~Santurkar, L.~Engstrom, A.~Turner, and A.~Madry, ``Robustness
  may be at odds with accuracy,'' \emph{arXiv preprint arXiv:1805.12152}, 2018.

\bibitem{schmidt2018adversarially}
L.~Schmidt, S.~Santurkar, D.~Tsipras, K.~Talwar, and A.~Madry, ``Adversarially
  robust generalization requires more data,'' in \emph{Advances in Neural
  Information Processing Systems (NIPS)}, 2018, pp. 5014--5026.

\bibitem{mnih2016asynchronous}
V.~Mnih, A.~P. Badia, M.~Mirza, A.~Graves, T.~Lillicrap, T.~Harley, D.~Silver,
  and K.~Kavukcuoglu, ``Asynchronous methods for deep reinforcement learning,''
  in \emph{International Conference on Machine Learning (ICML)}, 2016, pp.
  1928--1937.

\bibitem{bhatnagar2008incremental}
S.~Bhatnagar, M.~Ghavamzadeh, M.~Lee, and R.~S. Sutton, ``Incremental natural
  actor-critic algorithms,'' in \emph{Advances in Neural Information Processing
  Systems (NIPS)}, 2008, pp. 105--112.

\bibitem{sutton1998reinforcement}
R.~S. Sutton and A.~G. Barto, \emph{Reinforcement Learning: An
  Introduction}.\hskip 1em plus 0.5em minus 0.4em\relax MIT press Cambridge,
  1998, vol. 135.

\bibitem{kuutti2019end}
S.~Kuutti, R.~Bowden, H.~Joshi, R.~de~Temple, and S.~Fallah, ``End-to-end
  reinforcement learning for autonomous longitudinal control using advantage
  actor critic with temporal context,'' in \emph{2019 IEEE 22nd Intelligent
  Transportation Systems Conference (ITSC)}.\hskip 1em plus 0.5em minus
  0.4em\relax IEEE, 2019, pp. 2456--2462.

\bibitem{hochreiter1997long}
S.~Hochreiter and J.~Schmidhuber, ``Long short-term memory,'' \emph{Neural
  computation}, vol.~9, no.~8, pp. 1735--1780, 1997.

\bibitem{krizhevsky2010convolutional}
A.~Krizhevsky and G.~Hinton, ``Convolutional deep belief networks on
  cifar-10,'' \emph{Unpublished manuscript}, vol.~40, no.~7, 2010.

\bibitem{tieleman2012lecture}
T.~Tieleman and G.~Hinton, ``Lecture 6.5-rmsprop: Divide the gradient by a
  running average of its recent magnitude,'' \emph{COURSERA: Neural networks
  for machine learning}, vol.~4, no.~2, pp. 26--31, 2012.

\bibitem{IPG2017}
\BIBentryALTinterwordspacing
{IPG Automotive GmbH}, ``Carmaker: Virtual testing of automobiles and
  light-duty vehicles,'' 2017. [Online]. Available:
  \url{https://ipg-automotive.com/products-services/simulation-software/carmaker/}
\BIBentrySTDinterwordspacing

\bibitem{kuutti2019safe}
S.~Kuutti, R.~Bowden, H.~Joshi, R.~de~Temple, and S.~Fallah, ``Safe deep neural
  network-driven autonomous vehicles using software safety cages,'' in
  \emph{2019 20th International Conference on Intelligent Data Engineering and
  Automated Learning (IDEAL)}.\hskip 1em plus 0.5em minus 0.4em\relax Springer,
  2019, pp. 150--160.

\bibitem{reif2014brakes}
K.~Reif, ``Brakes, brake control and driver assistance systems,''
  \emph{Weisbaden, Germany, Springer Vieweg}, 2014.

\bibitem{henderson2018deep}
P.~Henderson, R.~Islam, P.~Bachman, J.~Pineau, D.~Precup, and D.~Meger, ``Deep
  reinforcement learning that matters,'' in \emph{Thirty-Second AAAI Conference
  on Artificial Intelligence}, 2018.

\bibitem{schulman2015high}
J.~Schulman, P.~Moritz, S.~Levine, M.~Jordan, and P.~Abbeel, ``High-dimensional
  continuous control using generalized advantage estimation,'' \emph{arXiv
  preprint arXiv:1506.02438}, 2015.

\bibitem{romoff2018reward}
J.~Romoff, P.~Henderson, A.~Pich{\'e}, V.~Francois-Lavet, and J.~Pineau,
  ``Reward estimation for variance reduction in deep reinforcement learning,''
  \emph{arXiv preprint arXiv:1805.03359}, 2018.

\bibitem{weaver2001optimal}
L.~Weaver and N.~Tao, ``The optimal reward baseline for gradient-based
  reinforcement learning,'' in \emph{Proceedings of the Seventeenth conference
  on Uncertainty in artificial intelligence}.\hskip 1em plus 0.5em minus
  0.4em\relax Morgan Kaufmann Publishers Inc., 2001, pp. 538--545.

\end{thebibliography}

\end{document}